\documentclass[a4paper,fleqn]{cas-sc}

\usepackage[authoryear]{natbib}

\usepackage{amsmath, bm}
\usepackage{algorithm}
\usepackage{comment}
\usepackage{lineno}
\usepackage{setspace}
\usepackage{subcaption}
\usepackage{siunitx}

\def\tsc#1{\csdef{#1}{\textsc{\lowercase{#1}}\xspace}}
\tsc{WGM}
\tsc{QE}

\newcommand{\etal}{\emph{et~al.}}
\newcommand{\secref}[1]{Section~\ref{#1}}
\newcommand{\figref}[1]{Figure~\ref{#1}}
\renewcommand{\tabref}[1]{Table~\ref{#1}}

\begin{document}

\let\WriteBookmarks\relax
\def\floatpagepagefraction{1}
\def\textpagefraction{.001}

\shorttitle{Convolutional ViTs for SSS Segmentation}    

\shortauthors{H. Rajani \etal }  

\title [mode = title]{A Convolutional Vision Transformer for Semantic Segmentation of Side-Scan Sonar Data}



%

\author[1]{Hayat Rajani}

\cormark[1]


\ead{hayat.rajani@udg.edu}


\credit{Conceptualization of this study, Methodology, Software, Data Curation, Writing - Original Draft}

\affiliation[1]{organization={Computer Vision and Robotics Research Institute (VICOROB), University of Girona},
            addressline={Campus Montilivi, Edifici P4}, 
            city={Girona},
            citysep={}, 
            postcode={17003}, 
            state={Catalonia},
            country={Spain}}
 
\author[1]{Nuno Gracias}
\ead{ngracias@silver.udg.edu}
\credit{Conceptualization of this study, Project Administration, Supervision, Writing - Review and Editing}

\author[1]{Rafael Garcia}
\ead{rafael.garcia@udg.edu}
\credit{Conceptualization of this study, Supervision, Validation, Writing - Review and Editing}


\cortext[1]{Corresponding author}




\begin{abstract}
Distinguishing among different marine benthic habitat characteristics is of key importance in a wide set of seabed operations ranging from installations of oil rigs to laying networks of cables and monitoring the impact of humans on marine ecosystems. The Side-Scan Sonar (SSS) is a widely used imaging sensor in this regard. It produces high-resolution seafloor maps by logging the intensities of sound waves reflected back from the seafloor. In this work, we leverage these acoustic intensity maps to produce pixel-wise categorization of different seafloor types. We propose a novel architecture adapted from the Vision Transformer (ViT) in an encoder-decoder framework. Further, in doing so, the applicability of ViTs is evaluated on smaller datasets. To overcome the lack of CNN-like inductive biases, thereby making ViTs more conducive to applications in low data regimes, we propose a novel feature extraction module to replace the Multi-layer Perceptron (MLP) block within transformer layers and a novel module to extract multiscale patch embeddings. A lightweight decoder is also proposed to complement this design in order to further boost multiscale feature extraction. With the modified architecture, we achieve state-of-the-art results and also meet real-time computational requirements. We make our code available at ~\url{https://github.com/hayatrajani/s3seg-vit}
\end{abstract}

\begin{keywords}
 seafloor segmentation \sep 
 side-scan sonar \sep 
 vision transformer \sep
 convolutional transformer \sep
 real-time \sep
\end{keywords}

\maketitle



\section{Introduction}
\label{sec:introduction}

High-resolution maps of the seafloor are a central tool for environmental monitoring. If these maps include topographical features and bottom types correctly identified and classified, they would become fundamental for the scientific and economic exploration of the oceans. The surveys for creating such maps are typically conducted from oceanographic vessels, using different types of acoustic sensors. Although the use of optical sensors for autonomous underwater exploration has seen several advances, they are severely affected by light attenuation and color shift caused by the variability in water conditions. This imposes limitations on the range of operation, restricting samples to be acquired only over small areas. The use of acoustic sensors, on the other hand, makes it possible to perceive the underwater environment even in zero-visibility conditions regardless of the depth, also allowing them to cover a much wider area in a single pass.

The Side Scan Sonar (SSS) is one such acoustic sensor that is very widely used in marine surveys. It is very easily adaptable to numerous types of sea vessels without the need for specific configurations and consumes low power, making it very economical and easy to deploy. As such, this work is aimed at semantic segmentation of acoustic images acquired from a SSS. Specifically, we deal with the problem of large-scale survey and exploration using an Autonomous Underwater Vehicle (AUV) to automate the process of mission planning and sensing, where online processing of such data is crucial.

Much of the prior work carried out in this area makes use of traditional image processing and pattern recognition approaches such as clustering strategies \citep{celik2011novel,yao2000unsupervised}, Markov Random Field (MRF) \citep{mignotte1999three, mignotte2000sonar} or active contouring \citep{lianantonakis2007sidescan}. These methods are based on hand-crafted features and either lack the efficiency to be used online or the capacity for generalization, among other issues.

In this paper, we propose to leverage the capabilities that Deep Neural Networks (DNNs) have showcased in pixel-wise labeling in recent years. In particular, we adopt Vision Transformers (ViTs) for the abovementioned task due to their ability to draw long-range associations among different regions of an image. The motivation behind using long-range associations comes from the fact that expert geophysicists often use global context to disambiguate among similarly looking classes. We believe that ViTs would enable the model to efficiently capture enough global context so as to make more informed decisions. This study, thus, also serves as a proof-of-concept of the feasibility of ViTs for tasks such as SSS segmentation. Given sufficient speed of computation, we can then use larger images (512x512) to capture more global information and further boost results. 

\begin{figure}
    \centering
    \includegraphics[width=0.8\columnwidth]{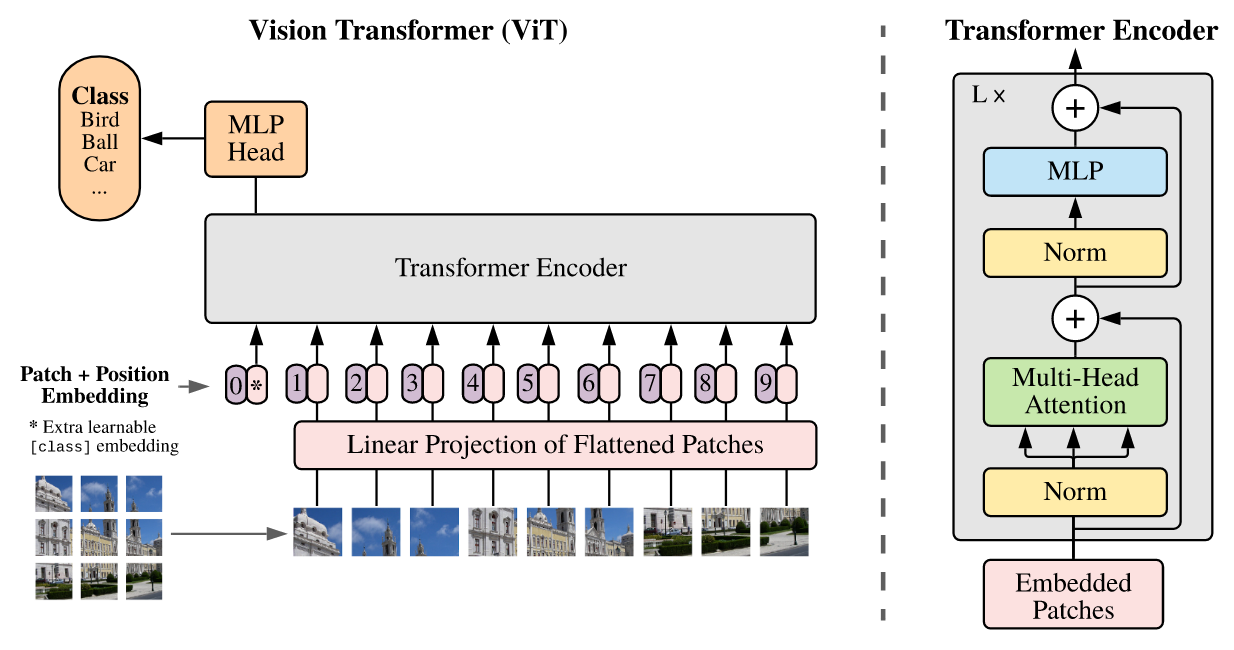}
    \caption{The original ViT architecture as proposed by \cite{dosovitskiy2021vit}}
    \label{fig:vit}
\end{figure}

The Transformer was originally developed as a sequence transduction model, for tasks such as machine translation, where it is essential for the model to formulate a thorough understanding of the language by capturing how the different components in a text interact with each other, the different semantics they might adopt in different contexts and the various syntactical patterns that might arise. \cite{vaswani2017attention} designed a computationally efficient mechanism, called the multi-head self-attention, to capture such global dependencies, which eventually led Transformer-based models to become the state-of-the-art in many Natural Language Processing (NLP) tasks. Later, \cite{dosovitskiy2021vit} transferred these principles to computer vision, giving rise to an architectural paradigm called the Vision Transformer, as depicted in \figref{fig:vit}.

Since the original Transformer was designed to operate on a sequence of 1D word embeddings, as opposed to the 2D or 3D images that vision models usually deal with, ViT breaks down the input image into a sequence of patches and linearly projects the flattened patches into an embedding space. This is essentially what allows the multi-head self-attention mechanism to ingest the $n$-D input and draw associations between different regions of the image, thereby capturing  global context. However, in doing so, the spatial correlations between patches are lost and the structure of the image is no longer preserved. This requires the use of positional encodings as a way to embed this structural information within the architecture. Unfortunately, positional encodings must be learnt from scratch as the model has no knowledge about the relative location of the patches to begin with. This increases the dependence of ViTs on large datasets, resulting in poor generalization otherwise. However, for domains such as marine robotics, where the data is typically very scarce, employing such architectures becomes infeasible, especially without any pre-training.

Convolutional Neural Networks (CNNs), on the other hand, have the grid-structure of the image built into their architectural design. This acts as a strong prior for the model resulting in properties such as shift invariance and equivariance. Moreover, although ViTs posses a global receptive field, convolutions tend to be more effective in extracting local features. We, therefore, propose two architectural modifications in an attempt to inject these characteristics into ViTs and to make them more suitable for applications that lack sufficiently large datasets.

\begin{itemize}
    \item We replace the linear patch embeddings at the beginning of each transformer stage with multiscale patch embeddings inspired in part by the design choices of Inception-v2 \citep{szegedy2016rethinking}.
    \item We replace the non-linear projections of the Multi-layer Perceptron (MLP) block within each transformer layer with a feature extraction block inspired by Ghost convolutions \citep{han2020ghostnet}.
\end{itemize}

The benefits are multifold. Not only does the use of convolutions within these modules introduce the notion of a grid-like structure, but posing them as stacked separable convolutions also reduces the number of parameters that the model needs to optimize. This further relaxes the need for position encodings while also enabling multiscale feature extraction. Consequently, this improves the capabilities of our modified architecture in capturing high-frequency details and generalizing well in the absence of huge datasets for training.

Furthermore, (hierarchical) ViT-based encoders, due to their ability to draw associations between different regions of an image, especially at multiple scales, tend to produce strong latent representations, which are readily suitable for the task of semantic segmentation. This permits the use of simple decoder designs without the need for computationally expensive modules. We model our decoder after the lightweight design proposed by \cite{xie2021segformer}, by supplementing it with auxiliary output blocks inspired by DeepLab’s Atrous Spatial Pyramid Pooling (ASPP) \citep{chen2018encoder}.

With our modified design, we surpass the results of previous state-of-the-art by a significant margin while also meeting the computational considerations for real-time implementation. Consequently, we demonstrate the applicability of ViTs for tasks such as semantic segmentation of the seafloor using SSS waterfalls, for which large datasets are seldom available. We believe that such hybrid ViT-based architectures have a large potential in underwater applications. To encourage further exploration of these approaches and to ease the reproducibility of our results, our code will be made available online at ~\url{https://github.com/hayatrajani/s3seg-vit}.

The remainder of this paper is organized as follows. \secref{sec:literature} presents a brief review of previous works on SSS segmentation using DNNs and related literature in the context of semantic segmentation using ViTs. Then, \secref{sec:methodology} provides thorough details of our proposed architecture. Next, \secref{sec:setup} presents an overview of the dataset and the experimental setup. \secref{sec:results} reports and visualizes our results. And finally, \secref{sec:conclusion} concludes this study by outlining the planned efforts.


\section{Related Work}
\label{sec:literature}

\subsection{CNNs for SSS segmentation}

The approach in this paper is, to the best of our knowledge, the first to demonstrate the applicability of ViTs to SSS segmentation. Nonetheless, a number of previous studies  have adopted CNNs and attention-based mechanisms for this task. \cite{wang2019rt} propose a U-Net \citep{ronneberger2015u} like encoder-decoder architecture for real-time SSS segmentation. They employ a two-way branching structure exploiting depth-wise separable convolutions in their encoder for efficiency and a combination of pooling indices and direct skip connections to feed the lost spatial information back into the decoder. \cite{wu2019ecnet}, on the other hand, focus on dealing with different sources of noise in SSS imagery and the issue of class imbalance. They employ residual blocks in their encoder and propose the use of side-output blocks, in addition to the typical encoder-decoder design, to leverage multi-level information from each encoder. Similarly, \cite{zhao2021dcnet} focus on dealing with different sources of noise in SSS imagery for real-time segmentation. Apart from carefully designing the encoder and decoder modules, they propose a novel DCblock employing dilated convolutions that sits between the encoder and the decoder to attain more context. Whereas, \cite{burguera2020line} target the problem of building a semantic map of the seafloor, specifically to search for loop candidates in a SLAM context. They propose an end to end framework in this regard while employing a lightweight encoder-decoder architecture for online multi-class SSS segmentation. However, \cite{wang2022fused} argue that such simple encoder-decoder designs are vulnerable to noise interference and only work well for SSS images with simple backgrounds. They propose an adaptive receptive field mechanism on the skip connections between the encoder and decoder branches to improve target shape fit. They further supplement the encoder branch with dynamic multiscale dilated convolution blocks to extract multiscale target features, and supplement the decoder branch with attention-based feature fusion blocks to better fuse global and local features while suppressing background noise. Furthermore, they propose a tree structure optimization module to refine the produced segmentation masks, thus, reducing the rate of misclassifications. \cite{wang2022fused} also propose a boundary loss based on structural similarity and weighted binary cross-entropy to improve classification along the contour of the target. However, we believe these enhancements to be specifically directed towards the problem of target segmentation, which is the main purpose of their work, as opposed to our objective of seafloor segmentation. Moreover, from the reported results, the architecture seems to have a very large number of parameters, and consequently a much lower inference speed. Therefore, we do not draw direct comparisons with this approach. \cite{yu2021side}, on the contrary, propose a novel approach for SSS segmentation by employing recurrent residual convolutions to capture global context followed by a self-guidance block for further refinement of results. The self-guidance block takes inspiration from the discriminator component of Generative Adversarial Networks (GANs) and serves to distinguish the generated segmentation mask from the ground truth. Although the authors claim a boost in segmentation results with their approach, the inference speed is significantly slow. Moreover, they mainly draw comparisons with models such as the U-Net \citep{ronneberger2015u} and SegNet \citep{badrinarayanan2017segnet}, which are over 15 times larger than our proposed architecture. We, therefore, do not include this approach in our comparative study either. \cite{yu2022dual}, take this approach further in another publication directed towards target segmentation. They propose a dual-branch framework comprising a segmentation branch and a refinement branch. The segmentation branch adopts a MobileNetv2 \citep{sandler2018mobilenetv2} backbone, additionally consisting of local attention and recurrent residual modules to dampen the effect of irrelevant features, thereby facilitating better emphasis on the target. This also addresses the overfitting caused by unbalanced datasets. The refinement branch further tunes this output with the help of holistic attention blocks for multi-level feature fusion. Both branches are further complemented by ASPP modules \citep{chen2018encoder} for enlarged receptive fields and better contextual understanding. However, this results in an architecture that is parameter-heavy, being almost 6 times larger than ours. Further, since this work is also intended for target segmentation, as opposed to our objective of seafloor segmentation, we do not draw direct comparisons with it.

\subsection{ViTs for Semantic Segmentation}

The vanilla ViT, as proposed by \cite{dosovitskiy2021vit}, yields low resolution feature maps of uniform scale, making it less desirable for dense prediction tasks such as semantic segmentation. To address this, \cite{wang2021pyramid} proposed as an alternative a hierarchical structure composed of a progressively shrinking pyramid capable of extracting features at multiple scales. This readily allowed ViTs to be plugged into standard dense prediction frameworks. They further apply spatial reduction to \textit{key} and \textit{value} embeddings before attention computation to handle the quadratic complexity associated with the traditional self-attention operation. Based on this hierarchical design, \cite{ren2022shunted} and \cite{yao2022wave} propose different flavours of spatially-reduced attention to preserve image details. Where the former downsamples the \textit{key} and \textit{value} embeddings with different rates for different attention heads, the latter employs discrete wavelet transforms. \cite{liu2021swin}, on the other hand, proposed a window-based self-attention scheme to reduce memory and computational costs. Attention is computed among tokens within non-overlapping local windows and the windows are shifted by a certain amount between consecutive layers to facilitate cross-window connections. This approach to self-attention computation was followed up by several other works \citep{huang2021shuffle, wang2021crossformer, dong2022cswin, wu2022pale} proposing different schemes of windowing and cross-flow of information among windows. However, these approaches are still restricted by the number of input tokens. To increase efficiency in processing high-resolution images, \cite{ali2021xcit} propose cross-covariance attention which computes self-attention among feature channels thereby making the computation linear in the number of tokens. \cite{koohpayegani2022sima} take this a step forward by replacing the softmax in self-attention with L1-normalization of \textit{key} and \textit{value} embeddings to further boost computational efficiency. In our proposed architecture, we use an adaptation of this approach, and draw comparisons with notable window-based and spatially-reduced self-attention mechanisms.

A related but independent line of work addresses the lack of CNN-like inductive biases in ViTs in order to improve their efficiency in capturing high-frequency details. There have been several studies \citep{dascoli2021convit, srinivas2021bottleneck, wu2021cvt, guo2022cmt, li2022contextual, ma2022mocovit, mehta2022mobilevit, si2022inception, zhang2022vitaev2} in this regard, which adopt one or a combination of diverse strategies such as introducing convolutions within the transformer blocks of ViTs, modifying the self-attention computation using convolutions, replacing the MLP blocks of ViTs with convolutions, or introducing transformer blocks within CNNs. The approach by \cite{guo2022cmt} is the closest to ours in that they also use a convolutional stem and adapt the MLP block of ViTs by introducing convolutions, apart from adopting convolutions inside their self-attention mechanism. However, in addition to inducing CNN-like inductive biases in the architecture, one of our primary objectives is to enhance the representability of objects of different scales, driving us to adopt distinct designs.

While the aforementioned studies focus on modifying the design of the ViT-based encoder counterpart of the framework, there have been a handful of efforts directed towards an efficient decoder design. For instance, \cite{cao2021swinu}, taking inspiration from U-Net \citep{ronneberger2015u}, propose a Swin Transformer \citep{liu2021swin} based symmetric upsampling decoder, and \cite{strudel2021segmenter} propose as a decoder a mask transformer that jointly processes patch and class embeddings. \cite{bousselham2021selfensemble}, on the other hand, propose an end-to-end trainable self-ensembling approach to leverage multi-scale features that are produced by different stages of the encoder, without the need of expensive feature fusion operations. However, \cite{xie2021segformer} propose a much simpler design, solely consisting of feature aggregating MLP blocks, that is capable of producing powerful representations while being computationally inexpensive. We model our decoder after an adaptation of this approach as detailed in \secref{sec:methodology:decoder}.


\section{Proposed Architecture}
\label{sec:methodology}

\begin{figure}
    \centering
    \includegraphics[width=0.7\columnwidth]{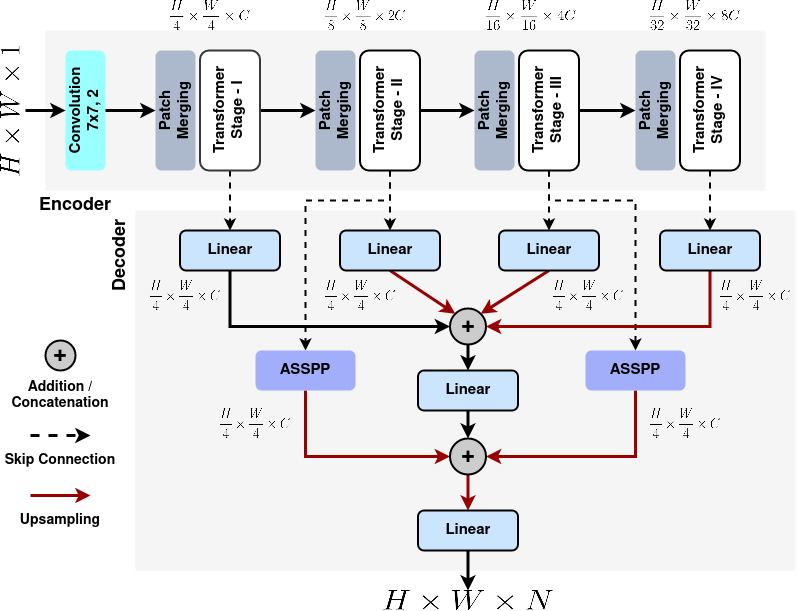}
    \caption{Overview of the proposed architecture. $H'$ and $W'$ refer to the spatial dimensions of the input. $C$ denotes the initial embedding dimension. $N$ denotes the number of classes.}
    \label{fig:s3seg}
\end{figure}

\figref{fig:s3seg} depicts an overview of the proposed architecture. It follows the general scheme of an encoder-decoder structure for semantic segmentation consisting of downsampling encoder blocks and upsampling decoder blocks with skip connections from the corresponding encoders. However, instead of building the decoder as a symmetric counterpart of the encoder, a much lighter-weight approach is adopted that is not only computationally more efficient but also yields better segmentation results, as discussed later in \secref{sec:results}. The following two subsections describe the encoder and decoder modules in further detail.

\subsection{A Multiscale Convolutional ViT Encoder}
\label{sec:methodology:encoder}

The encoder adopts a modified version of a conventional 4-stage hierarchical ViT that begins with an initial patch size of $4\times4$ pixels, projected onto a $C$-dimensional embedding space. We set the initial length of embeddings to 24. For an input of size $H \times W$ pixels, this results in a sequence of $\,(H\cdot W)$ \textfractionsolidus $\,16\,$
24-dimensional patch embeddings. Between two successive transformer stages, patches in non-overlapping $2\times2$ neighborhoods are merged together while the length of their embeddings is doubled. This essentially downsamples each patch by a factor of 2 and, consequently, reduces the sequence length by a factor of 4. As a result, each transformer stage operates on a different scale, thereby making it possible for the encoder to construct a feature pyramid comparable to that of traditional CNN backbones.

Traditionally, the patch embedding and merging modules are composed of reshaping, flattening and linear projection operations. Our approach, on the other hand, leverages the local feature extraction capabilities of convolutions. We begin by applying a $7\times7$ convolution with a stride of 2 to the input image, resulting in a feature representation of size $\frac{H}{2}\times\frac{W}{2}\times12$. 
Before each transformer stage, we then place a patch merging module that downsamples the input by a factor of 2 and doubles the number of feature channels. The produced feature representation is subsequently flattened along the spatial dimensions to make it suitable to be processed by the corresponding transformer stage. Thus, each transformer stage, $i \in \{1,2,3,4\}$, operates on an input of size $\frac{H \cdot W}{2^{i+1}}\times12$$\cdot2^i$.

\begin{figure}
    \centering
    \includegraphics[width=0.8\columnwidth]{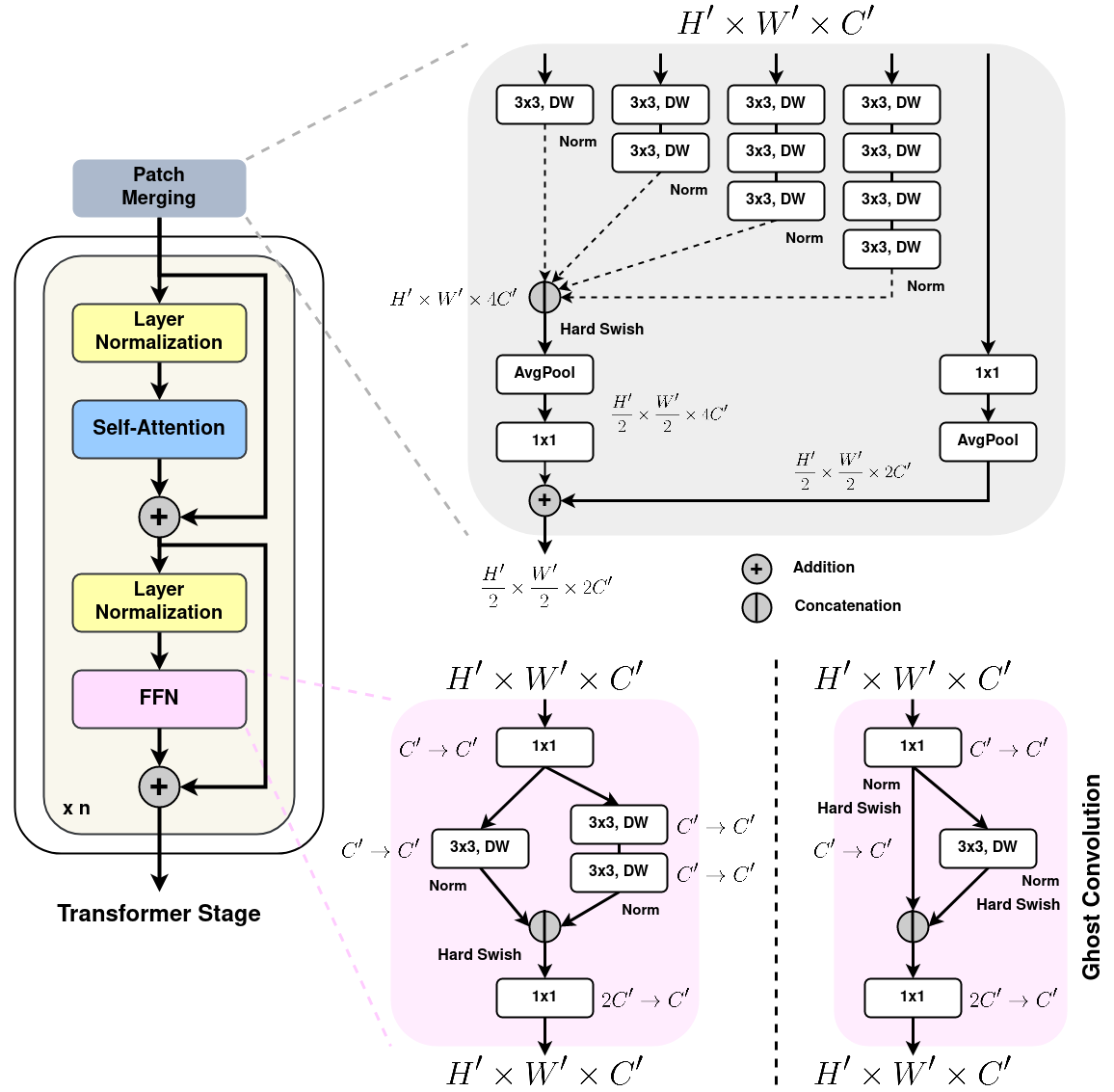}
    \caption{Overview of a transformer stage together with illustrations of the patch merging module, the proposed FFN and ghost convolutions. $H'$ and $W'$ refer to the spatial dimensions of the input. $C'$ denotes the number of input channels. $DW$ denotes a depthwise convolution.}
    \label{fig:Tstage}
\end{figure}

The patch merging module, as illustrated in \figref{fig:Tstage}, was in part inspired by the design of Inception-v2 \citep{szegedy2016rethinking}. It consists of four parallel branches of stacked depthwise convolutions of different receptive fields with appropriate padding to maintain spatial dimensions. Apart from introducing convolutional priors, the main rationale behind the patch merging module was to be able to adequately represent objects of different scales such as small pebbles to large boulders. The parallel convolutional branches enable feature extraction over various spatial footprints to generate multi-scale patch embeddings. Such a design further complements self-attention to draw finer global associations. Further, employing depthwise convolutions instead of full convolutions and factorizing convolutions with larger kernels by a stack of $3\times3$ convolutions, significantly saves on parameters while maintaining the effective receptive field. Each depthwise convolution operation is also followed by group normalization \citep{wu2018group} with the number of groups set to 1. We then apply average pooling, preceded by a Hard Swish non-linearity \citep{howard2019searching}, to downsample the generated feature representations by a factor of 2 followed by a pointwise convolution to project the aggregated multi-scale representations to twice the length of the input embeddings. Moreover, we introduce a residual connection, composed of a pointwise convolution and average pooling, for stability. The overall parameter count still remains lower as compared to employing a single full $3\times3$ convolution with a stride of 2 for patch merging, as proposed by \cite{wu2021cvt} and \cite{dong2022cswin}, in lieu of linear projections.

\figref{fig:Tstage} also illustrates the transformer stage in more detail. Each transformer stage consists of $L \in \{3,6,12,3\}$ transformer layers. Each transformer layer, as in the original ViT design \citep{dosovitskiy2021vit}, is composed of a self-attention module and an MLP block, each preceded by layer normalization \citep{ba2016layer} and accompanied by a residual connection \citep{he2016deep}. Traditionally, the MLP block comprises two linear projections separated by a GELU non-linearity. Since na\"ively replacing the linear projections by $3\times3$ convolutions increases the overall parameter count 9-fold, we instead adopted ghost convolutions as conceived by \cite{han2020ghostnet}. They try to replicate the redundancy in feature maps generated by full convolutions through cheap linear operations. Specifically, a full convolution is split into two parts: a pointwise convolution in order to generate the primary feature maps followed by a $3\times3$ depthwise convolution (portrayed as a cheap linear operation) in order to generate secondary "ghost" features that add redundancy. We later extended this design by applying an additional $5\times5$ depthwise convolution, implemented as a stack of two $3\times3$ depthwise convolutions, to the primary features before concatenating them with the ghost features. We use this extended ghost convolution to replace the first linear projection in the original MLP. Again, after each convolution operation, we employ group normalization with a group size of 1. We also employ a Hard Swish non-linearity before the final linear projection. The resultant module not only introduces image-specific priors into the design but also reduces the parameter count when compared to a corresponding MLP block with an expansion ratio of 2.

Finally, due to the linear complexity in the number of patches, we use the attention mechanism as proposed by \cite{ali2021xcit}. To further reduce computational cost, we remove the expensive softmax-based normalization of the attention matrix and, instead, L1-normalize the \textit{key} and \textit{query} embeddings before computing the attention scores as proposed by \cite{koohpayegani2022sima}. To avoid ambiguity in comparison, we term this modified attention mechanism \textbf{\emph{SimXCA}}, a contraction of the names of the two referenced approaches. We set the number of attention heads to $\{2,4,8,16\}$ for the four transformer stages respectively, and we do not use any kind of positional encoding.

\subsection{A SegFormer-ASSPP Decoder}
\label{sec:methodology:decoder}

Classical CNN-based approaches to semantic segmentation, such as U-Net \citep{ronneberger2015u} and SegNet \citep{badrinarayanan2017segnet}, design the decoder as a symmetric counterpart to the encoder together with skip connections in order to add the lost spatial information back from the multi-resolution feature pyramid of the encoder. However, we observed that adopting such a symmetric design for ViT-based encoder-decoder frameworks yields quite poor results. This can be particularly attributed to the dependence of ViTs on large datasets to overcome their inherent lack of CNN-like inductive biases, which is further amplified by the symmetric decoder. Despite our modifications to induce CNN-like inductive biases in the design, we postulate that such complex architectures are not really necessary for the decoder. The encoder, due to its ability to draw associations between different regions of the input image and extract multi-scale local features, already tends to produce strong latent representations that are readily suitable for the task of semantic segmentation. This permits the use of simpler designs for the decoder, which also drastically reduces the number of trainable parameters. As such, we base our decoder after the design proposed by \cite{xie2021segformer}.

First, feature representations from each encoder stage are linearly projected to the initial embedding dimension, $C$, and are bilinearly upsampled to match the size of the initial patch embedding, $\frac{H}{4}\times\frac{W}{4}$. The projected representations are then fused together by addition, instead of concatenation as proposed by \cite{xie2021segformer}. This is followed by another linear projection to the initial embedding dimension. This fused representation then undergoes a final linear projection to an $N$-dimensional space, where $N$ is the number of classes, in order to predict the segmentation mask, which is subsequently upsampled by a factor of 4 to match the input image resolution.

However, we postulate that such a simple feature aggregating decoder is unable to completely leverage the multi-scale representations generated by the encoder, especially for small-scale objects. Therefore, contrary to \cite{xie2021segformer}, we also employ two auxiliary blocks, based on the ASPP module \citep{chen2018encoder}, that operate on the feature representations from the second and the third encoder stages. The ASPP module can effectively enlarge the receptive field and incorporate multi-scale context, which also compensates for the lack of explicit modeling of global associations in the decoder, thereby further reducing the rate of misclassification. Since the bulk of the transformer blocks lie within the third stage of the encoder, the so produced feature representations are rich-enough for ASPP-decoding. Also, the spatial resolution is adequately high to not suffer significantly from the loss of information caused by downsampling. To further sharpen the results, we also adopt the feature representations from the second encoder stage for ASPP-decoding due to the relatively higher spatial resolution.

\begin{figure}
    \centering
    \includegraphics[width=0.4\columnwidth]{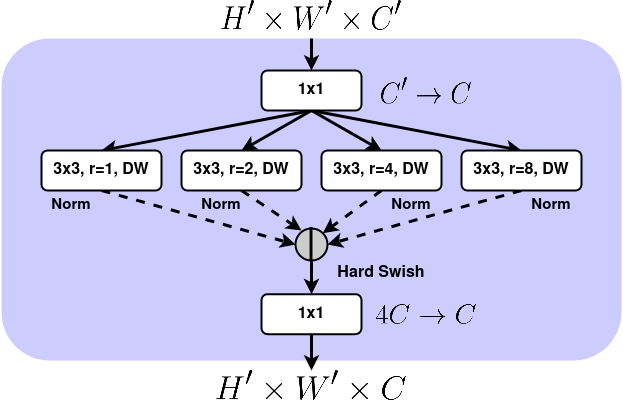}
    \caption{Overview of the modified ASPP module. $H'$ and $W'$ refer to the spatial dimensions of the input. $C'$ denotes the number of input channels. $C$ denotes the initial embedding dimension. $DW$ denotes a depthwise convolution.}
    \label{fig:asspp}
\end{figure}

Our modified version of the ASPP module is illustrated in \figref{fig:asspp}. First, we project the input feature representations onto the initial embedding dimension, $C$. The projected representation is then passed through the four parallel atrous convolution branches with dilation rates $r \in \{1,2,4,8\}$, and a kernel size of 3. Each atrous convolution is implemented as a depthwise convolution and is followed by group normalization \citep{wu2018group} with the number of groups set to 1. The feature representations from each atrous branch are then concatenated together and projected to the initial embedding dimension. This projection is preceded by a Hard Swish non-linearity \citep{howard2019searching}. Finally, the aggregated representation from the two auxiliary blocks is upsampled to match the size of the initial patch embeddings. These upsampled representations are then concatenated with the previously fused representation from all four encoder stages, before predicting the final segmentation mask.


\section{Experimental Setup}
\label{sec:setup}

\subsection{Dataset}
\label{sec:setup:data}

The datasets used in the course of this work were acquired with a Klein 3000 Side Scan Sonar during various surveys in the Balearic Sea. Approximately \SI{52}{\kilo\meter} of coastal area was surveyed at an altitude varying from 4 to 21 meters. Four categories of sediments were identified, namely Sand Ripples, Rocks, Maerl and Fine Sediments (such as silt and mud) covering approximately \SI{50.60}{\percent}, \SI{13.90}{\percent}, \SI{12.06}{\percent} and \SI{23.44}{\percent} of the total area respectively.

\begin{figure}[t]
    \centering
    \includegraphics[width=0.8\columnwidth]{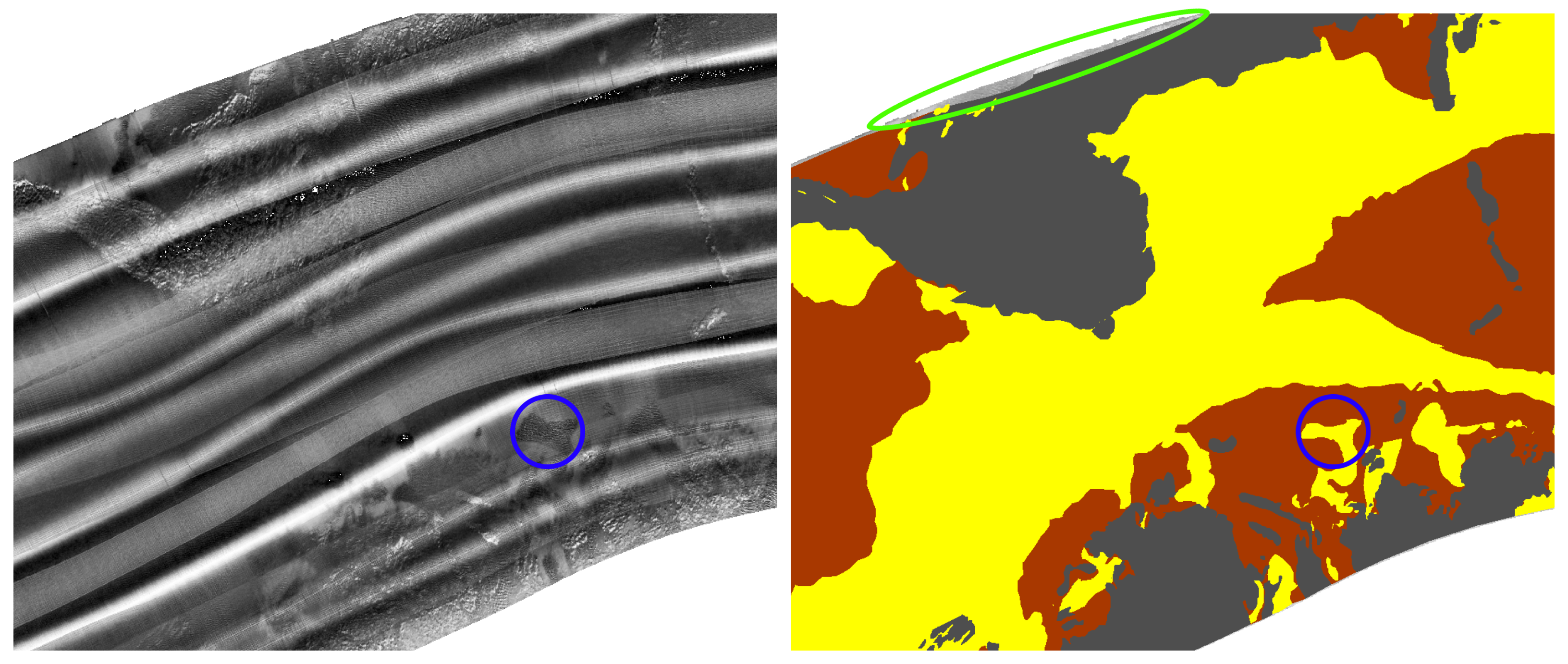}
    \caption{SSS mosaic (left) and the corresponding interpretation (right). Note that in some areas the annotations are not accurate (as marked in blue) and may also be missing for certain areas (as marked in green).}
    \label{fig:mosaic}
\end{figure}

The raw SSS waterfalls were recorded in the eXtended Triton Format (XTF) and subsequently processed using SonarWiz for mosaicing. The mosaiced SSS waterfalls were then georeferenced and annotated by two expert geophysicists using ArcGIS. \figref{fig:mosaic} depicts an example of a portion of the SSS mosaic and the corresponding annotation. We further processed the raw SSS waterfalls for blind zone removal and slat range correction. The available navigation data was then used to geocode each bin of the waterfall. This allowed us to fetch the corresponding annotations from the ArcGIS interpretations and automatically generate the ground truth for the SSS waterfalls. However, since the annotations were fetched from SSS mosaics while the ground truth was being generated for SSS waterfalls, misalignments in the mosaic may result in slight pixel-wise errors in the ground truth. Moreover, the annotations were made on a much coarser resolution than the SSS waterfalls and also suffered from human error resulting in ambiguous inter-class boundaries, missing labels and skewed or misaligned annotations for certain areas. However, despite the noisy ground truth, our model is able to generalize quite well, as discussed in \secref{sec:results}.

\begin{figure}[t]
    \centering
    \begin{tabular}{ccc}
        \subcaptionbox{}{\includegraphics[width=0.2\columnwidth]{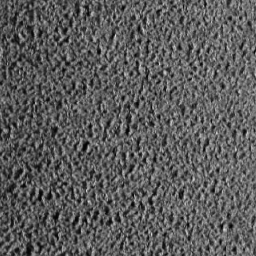}} &
        \subcaptionbox{}{\includegraphics[width=0.2\columnwidth]{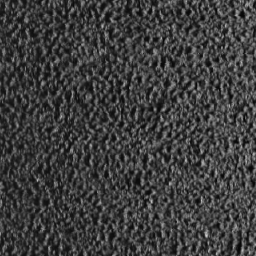}} &
        \subcaptionbox{}{\includegraphics[width=0.2\columnwidth]{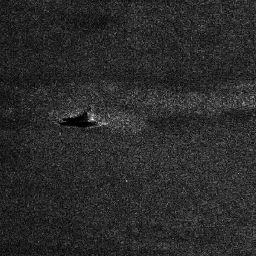}} \\
        \subcaptionbox{}{\includegraphics[width=0.2\columnwidth]{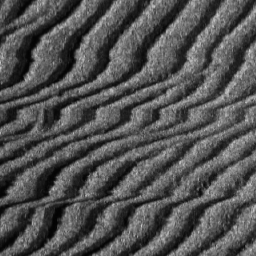}} &
        \subcaptionbox{}{\includegraphics[width=0.2\columnwidth]{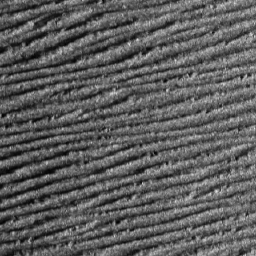}} &
        \subcaptionbox{}{\includegraphics[width=0.2\columnwidth]{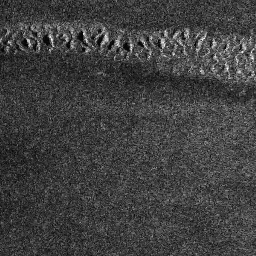}} \\
        \subcaptionbox{}{\includegraphics[width=0.2\columnwidth]{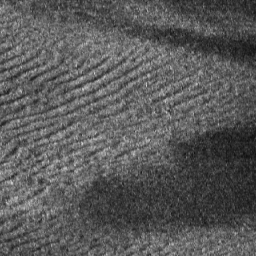}} &
        \subcaptionbox{}{\includegraphics[width=0.2\columnwidth]{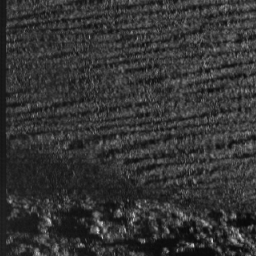}} &
        \subcaptionbox{}{\includegraphics[width=0.2\columnwidth]{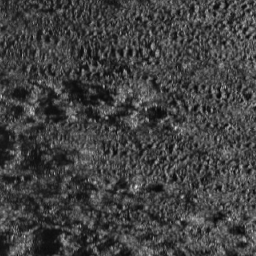}} \\
        \subcaptionbox{}{\includegraphics[width=0.2\columnwidth]{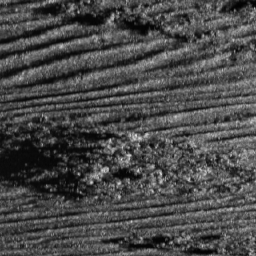}} &
        \subcaptionbox{}{\includegraphics[width=0.2\columnwidth]{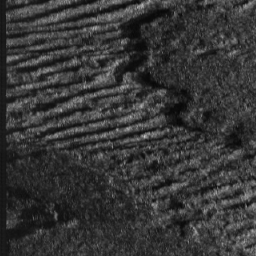}} &
        \subcaptionbox{}{\includegraphics[width=0.2\columnwidth]{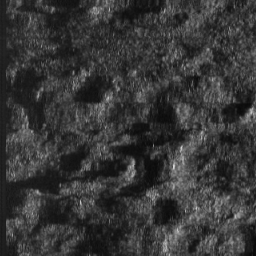}} 
    \end{tabular}
    \caption{Dataset overview: (a,b) gravel; (c) fine sediment with a chunk of rock; (d, e) rippled sand; (f) fine sediment with a strip of rippled sand; (g) fine sediment with rippled sand; (h) rocks, fine sediment and rippled sand; (i) rocks and gavel; (j, k) rocks and rippled sand; (l) rocks}
    \label{fig:dataset}
\end{figure}

The waterfalls and the corresponding ground truth were then partitioned in batches of 256 lines to generate images of size $256\times256$ with a 128 pixel-overlap along-track and across-track. This resulted in a total of 47,420 images, divided by an 80-\SI{20}{\percent} split to form the training and validation sets respectively. Further, another set of images, equivalent to about \SI{5}{\percent} of the training set, was generated from a separate non-overlapping transect with similar class distribution as the training set to form a test set of 1800 additional images. The noisy ground truth of these test images was then manually corrected so as to be able to produce accurate metrics for evaluation. \figref{fig:dataset} illustrates some examples of the images and the different seabed types contained in the dataset.

\subsection{Training and Evaluation}
\label{sec:setup:code}

All our models were trained on an NVIDIA A100 Tensor Core GPU for 100 epochs with a batch size of 64. We utilized the AdamW optimizer with a weight decay of $1e^{-2}$ and learning rate of $6e^{-5}$, decayed using a polynomial learning rate scheduler with a warm-up of 3 epochs. Weighted Cross Entropy was set as the loss function. We also adopted standard data augmentation techniques such as random rotation, random resized crop, random horizontal and vertical flip, random variations in contrast and/or sharpening and Gaussian blur. The models were implemented in PyTorch 1.11.0 and Python 3.8.10. The source code with all hyper-parameter configurations and pre-trained models will be made available at ~\url{https://github.com/hayatrajani/s3seg-vit}.

All trained models were then evaluated on a standard laptop equipped with an NVIDIA GeForce GTX 1650 Mobile GPU and an Intel Core i5-9300H CPU operating at \SI{2.40}{\giga\hertz} running Ubuntu 20.04.5, Python 3.8.10 and PyTorch 1.9.0+cu111. We report model performance in terms of mean Intersection over Union (mIoU) and inference speed in number of images processed per second (FPS).

\section{Experiments and Results}
\label{sec:results}

\subsection{Comparisons with state-of-the-art CNNs}

The CNN-based frameworks RTSeg \citep{wang2019rt}, ECNet \citep{wu2019ecnet} and DCNet \citep{zhao2021dcnet} were adopted as baselines for primary comparison. We use our own implementation of these architectures since the authors' code was not made publicly available. All the architectural configurations were maintained as suggested in the original publications. However, we noticed certain inconsistencies in the architectures for RTSeg and DCNet. The former had a deviation of about \SI{0.24}{M} between the number of parameters computed from the description of the architecture and that reported by \cite{wang2019rt}. Whereas in the latter, \cite{zhao2021dcnet} do not specify the output feature dimensions for convolutions in their DCblock. We tried to set the parameters that we considered best to replicate the architectures as closely as possible. However, the results might deviate from the original implementations of the respective authors.

\begin{table}[width=.9\columnwidth,cols=5,pos=t]
  \caption{Comparison between our proposed architecture and different state-of-the-art CNNs for SSS Segmentation. \\ Best results are indicated in red. Second best results are indicated in blue.}
  \begin{tabular*}{\tblwidth}{LCCCC}
  \toprule
    \textbf{Method} & \textbf{mIoU (\%)} & \textbf{Parameters (M)} & \textbf{FPS} & \textbf{Model Size (MB)}\\
    \midrule
    ECNet & 62.68 & 4.67 & 70 & 18.8 \\
    DCNet & 78.02 & 0.09 & 260 & \textcolor{red}{0.4} \\
    RTSeg & 77.75 & 0.74 & 190 & 3.1 \\
    \midrule
    Ours$^{\bm{\dagger}}$ & 81.24 & \textcolor{red}{0.08} & \textcolor{red}{525} & \textcolor{blue}{0.47} \\
    Ours$^{\bm{\dagger\dagger}}$ & 84.65 & \textcolor{blue}{0.26} & \textcolor{blue}{311} & 1.2 \\
    Ours$^{\bm{\ddagger}}$ & \textcolor{blue}{88.58} & 0.97 & 188 & 4.1 \\
    \midrule
    Ours & \textcolor{red}{89.54} & 1.91 & 112 & 7.9 \\
    \bottomrule
    \label{tab:cnn}
  \end{tabular*}
\end{table}

\tabref{tab:cnn} presents a comparison of our proposed architecture with the above-mentioned approaches. Although our architecture has a significantly larger number of parameters than DCNet and RTSeg, the gain in mIoU is also significantly higher. Further, given that the ping rate of our SSS is 20 per second, it takes about \SI{12.8}{\second} to collect 256 swaths. Since each swath has about 1024 bins per side (port and starboard) and we generate images of size $256\times256$ with a 128 pixel-overlap, the minimum processing speed required for real-time segmentation is 14 images per \SI{12.8}{\second}, which equates to \SI{1.01} frames per second. This is including the bins corresponding to the water column which we however do not take into account for segmentation. Even considering the overhead for I/O operations, pre-processing and stitching the segmented images back together into a coherent waterfall, our architecture is readily suitable for real-time segmentation.

In order to ensure a more fair comparison with the aforementioned approaches, \tabref{tab:cnn} also presents the results of three variants of our proposed architecture with a significant reduction in parameters to match those of RTSeg and DCNet. These variants are indicated as Ours$^{\bm{\dagger}}$, Ours$^{\bm{\dagger\dagger}}$ and Ours$^{\bm{\ddagger}}$. We reduce the lengths of the initial embeddings for the first two variants down to 8 and 12 respectively. We also reduce the number of attention heads down to $\{1,2,4,8\}$ for each of their respective transformer stages. Furthermore, for the first variant, we set the number of layers in each transformer stage to $\{1,1,3,1\}$, whereas for the second and third variants, we set the number of layers in each transformer stage to $\{1,3,7,1\}$. All other configurations remain the same. Despite the significant reduction in parameters, our architectures perform substantially better. With this, we establish a new state-of-the-art for SSS segmentation.

\subsection{Feasibility of ViTs for SSS Segmentation}

The secondary objective of our study is to evaluate the feasibility of ViTs for applications such as SSS segmentation, which typically lack sufficiently large datasets. We therefore draw comparisons of our modified architecture with certain notable self-attention mechanisms in the vanilla hierarchical ViT setting, as presented in \tabref{tab:vit}.

\begin{table}[width=.9\columnwidth,cols=5,pos=t]
  \caption{Comparison among different self-attention mechanisms for SSS Segmentation. \\ Best results are indicated in bold.}
  \begin{tabular*}{\tblwidth}{LCCCC}
  \toprule
    \textbf{Method} & \textbf{mIoU (\%)} & \textbf{Parameters (M)} & \textbf{FPS} & \textbf{Model Size (MB)}\\
    \midrule
    Swin & 84.94 & 2.35 & 160 & 10.3 \\
    CSWin & 84.95 & 2.12 & 146 & 8.7 \\
    LSDA & 84.96 & 2.12 & 186 & 8.7 \\
    LMHSA & 81.31 & 2.15 & 176 & \textbf{8.3} \\
    Wavelet & 85.33 & 4.37 & 152 & 17.3 \\
    SimXCA & 85.97 & \textbf{2.10} & \textbf{204} & 8.5 \\
    \midrule
    SimXCA$^{\bm{\dagger}}$ & \textbf{86.23} & 2.14 & 186 & 8.7 \\
    \bottomrule
    \label{tab:vit}
  \end{tabular*}
\end{table}

Specifically, to switch back to a vanilla hierarchical ViT, we discard all our architectural modifications and simply employ the original MLP block within each transformer layer and use $3\times3$ convolutions with a stride of 2 for patch merging. Furthermore, we adopt the decoder as proposed by \cite{xie2021segformer} in its original form, without our modified ASPP module. We then train different architectures by replacing the self-attention modules in each transformer layer with the self-attention mechanism proposed by \cite{liu2021swin} (Swin), \cite{dong2022cswin} (CSWin), \cite{wang2021crossformer} (LSDA), \cite{yao2022wave} (Wavelets block) and \cite{guo2022cmt} (LMHSA block). Since, the approach to self-attention proposed by \cite{wang2021crossformer} works by alternately applying long-distance attention and short-distance attention in different layers, it suggests the use of an even number of transformer layers for each transformer stage. Therefore, in order to ensure a fair comparison, we set the number of transformer layers $L$ to $\{2,4,16,2\}$ for the four encoder stages for all flavours of attention mechanisms, unless stated otherwise. Also, we do not include positional encodings when employing the Wavelets block, the LMHSA block or SimXCA as the attention mechanism, as suggested by the respective works. All other architectural configurations remain the same. Additional hyper-parameter configurations specific to each self-attention mechanism, are as given below:

\begin{itemize}
    \item \textbf{LSDA:} Group size, $G = 8$
    \item \textbf{SWin:} Window size, $M = 8$
    \item \textbf{CSWin:} Stripe Width for each encoder stage, $sw = \{1,2,8,8\}$
    \item \textbf{LMHSA block:} Spatial reduction scale of \textit{key} and \textit{value} embeddings for each encoder stage, $s = \{8,4,2,1\}$
    \item \textbf{Wavelet block:} Spatial reduction scale of \textit{key} and \textit{value} embeddings for each encoder stage, $s = \{4,2,1,1\}$
\end{itemize}

The LMHSA block generates quite poor segmentation masks due to the loss of information that results from spatial downsampling of \textit{key} and \textit{query} embeddings. The Wavelet block significantly recovers this drop in quality by leveraging wavelet transforms that allow invertible downsampling. However, this approach turns out to be too parameter-heavy and also quite computationally expensive. Although with a slightly lower mIoU, the window-based self-attention mechanisms are quite efficient. The variability in their performance stems from the adopted windowing mechanism, where LSDA performs the best due to the alternating structure of local and global attention that results in self-attention computation between fewer tokens than Swin or CSWin. However, window-based self-attention mechanisms are still a compromise to global self-attention. SimXCA, on the other hand, gives the best results owing to its transposed self-attention computation and absence of Softmax-based normalization. We also observed that redistributing the number of transformer layers $L$ to $\{3,6,12,3\}$ for SimXCA produces slightly better segmentation masks, while keeping the total number of transformer layers the same. We indicate this as SimXCA$^{\bm{\dagger}}$ in \tabref{tab:vit}.

\begin{table}[width=.9\columnwidth,cols=5,pos=t]
  \caption{Ablations of our proposed architecture.}
  \begin{tabular*}{\tblwidth}{LCCCC}
  \toprule
    \textbf{Method} & \textbf{mIoU (\%)} & \textbf{Parameters (M)} & \textbf{FPS} & \textbf{Model Size (MB)}\\
    \midrule
    SimXCA & 86.23 & 2.14 & 186 & 8.7 \\
    $\bm{-}$ MLP & 88.18 & 1.98 & 136 & 8.1 \\
    $\bm{+}$ Multimerge & 89.28 & 1.90 & 116 & 7.9 \\
    $\bm{+}$ ASSPP & 89.54 & 1.91 & 112 & 7.9 \\
    \bottomrule
    \label{tab:ablation}
  \end{tabular*}
\end{table}

In \tabref{tab:ablation}, on the other hand, we ablate our modified architecture to present the individual performance gains from each of our architectural modifications. The top row represents SimXCA in the vanilla hierarchical ViT setting with the number of transformer layers $L$ to $\{3,6,12,3\}$. We then replace the MLP block within each transformer layer with our modified feature extraction block, denoted as "- MLP". Next, we replace the patch merging module in the resultant architecture with our proposed multiscale patch merging module, denoted as "+ Multimerge". Finally, we add the proposed ASSPP block to the decoder design, completing our modified architecture.

\begin{figure}[t]
    \centering
    \begin{tabular}{cccc}
        \subcaptionbox*{}{\includegraphics[width=0.2\columnwidth]{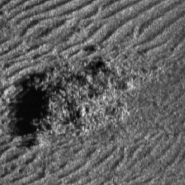}} &
        \subcaptionbox*{}{\includegraphics[width=0.2\columnwidth]{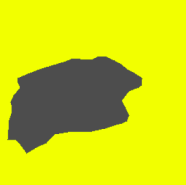}} &
        \subcaptionbox*{}{\includegraphics[width=0.2\columnwidth]{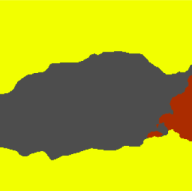}} &
        \subcaptionbox*{}{\includegraphics[width=0.2\columnwidth]{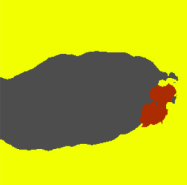}} \\
        \subcaptionbox*{}{\includegraphics[width=0.2\columnwidth]{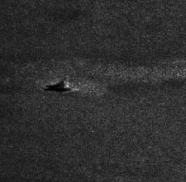}} &
        \subcaptionbox*{}{\includegraphics[width=0.2\columnwidth]{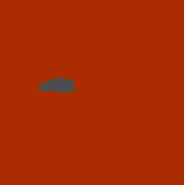}} &
        \subcaptionbox*{}{\includegraphics[width=0.2\columnwidth]{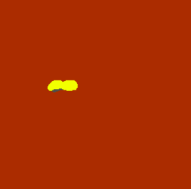}} &
        \subcaptionbox*{}{\includegraphics[width=0.2\columnwidth]{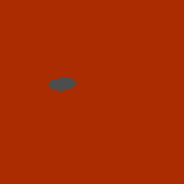}} \\
        \subcaptionbox*{}{\includegraphics[width=0.2\columnwidth]{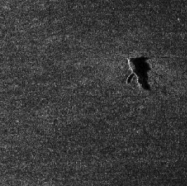}} &
        \subcaptionbox*{}{\includegraphics[width=0.2\columnwidth]{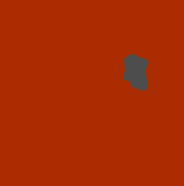}} &
        \subcaptionbox*{}{\includegraphics[width=0.2\columnwidth]{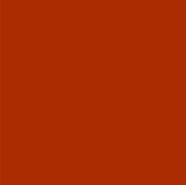}} &
        \subcaptionbox*{}{\includegraphics[width=0.2\columnwidth]{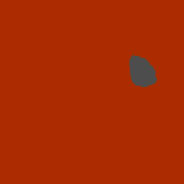}} \\
        \subcaptionbox*{}{\includegraphics[width=0.2\columnwidth]{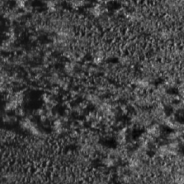}} &
        \subcaptionbox*{}{\includegraphics[width=0.2\columnwidth]{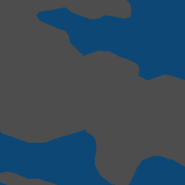}} &
        \subcaptionbox*{}{\includegraphics[width=0.2\columnwidth]{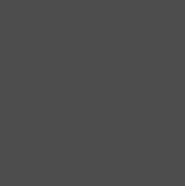}} &
        \subcaptionbox*{}{\includegraphics[width=0.2\columnwidth]{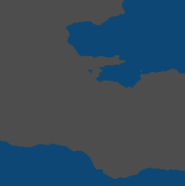}} \\
        \subcaptionbox*{Input}{\includegraphics[width=0.2\columnwidth]{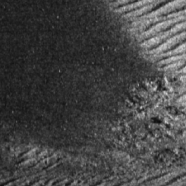}} &
        \subcaptionbox*{Ground Truth \\ (noisy)}{\includegraphics[width=0.2\columnwidth]{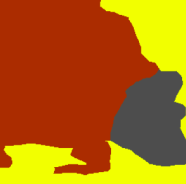}} &
        \subcaptionbox*{Vanilla}{\includegraphics[width=0.2\columnwidth]{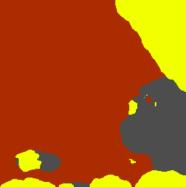}} &
        \subcaptionbox*{Ours}{\includegraphics[width=0.2\columnwidth]{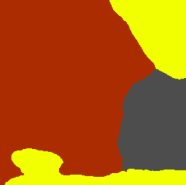}}
    \end{tabular}
    \caption{Visualization of segmentation masks.}
    \label{fig:viz-main-a}
\end{figure}

\begin{figure}[t]
    \centering
    \begin{tabular}{cccc}
        \subcaptionbox*{}{\includegraphics[width=0.2\columnwidth]{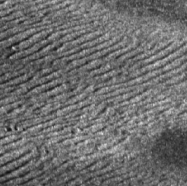}} &
        \subcaptionbox*{}{\includegraphics[width=0.2\columnwidth]{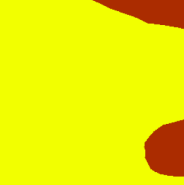}} &
        \subcaptionbox*{}{\includegraphics[width=0.2\columnwidth]{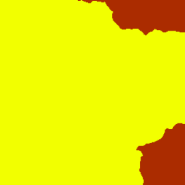}} &
        \subcaptionbox*{}{\includegraphics[width=0.2\columnwidth]{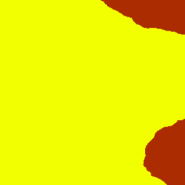}} \\
        \subcaptionbox*{}{\includegraphics[width=0.2\columnwidth]{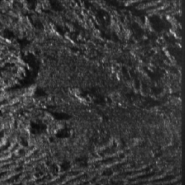}} &
        \subcaptionbox*{}{\includegraphics[width=0.2\columnwidth]{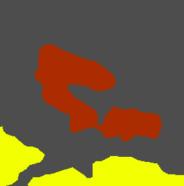}} &
        \subcaptionbox*{}{\includegraphics[width=0.2\columnwidth]{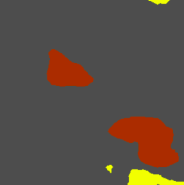}} &
        \subcaptionbox*{}{\includegraphics[width=0.2\columnwidth]{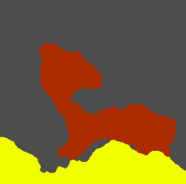}} \\
        \subcaptionbox*{}{\includegraphics[width=0.2\columnwidth]{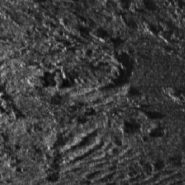}} &
        \subcaptionbox*{}{\includegraphics[width=0.2\columnwidth]{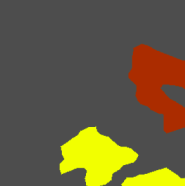}} &
        \subcaptionbox*{}{\includegraphics[width=0.2\columnwidth]{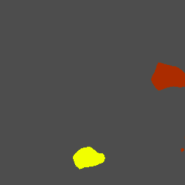}} &
        \subcaptionbox*{}{\includegraphics[width=0.2\columnwidth]{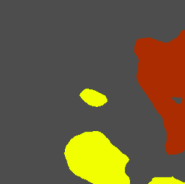}} \\
        \subcaptionbox*{}{\includegraphics[width=0.2\columnwidth]{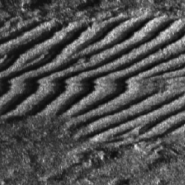}} &
        \subcaptionbox*{}{\includegraphics[width=0.2\columnwidth]{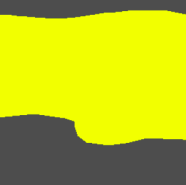}} &
        \subcaptionbox*{}{\includegraphics[width=0.2\columnwidth]{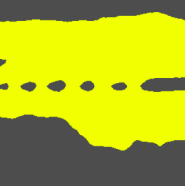}} &
        \subcaptionbox*{}{\includegraphics[width=0.2\columnwidth]{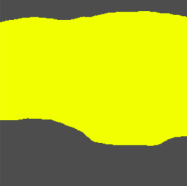}} \\
        \subcaptionbox*{Input}{\includegraphics[width=0.2\columnwidth]{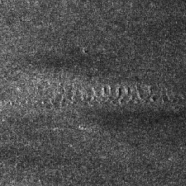}} &
        \subcaptionbox*{Ground Truth \\ (noisy)}{\includegraphics[width=0.2\columnwidth]{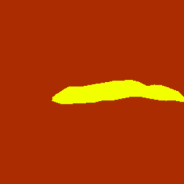}} &
        \subcaptionbox*{Vanilla}{\includegraphics[width=0.2\columnwidth]{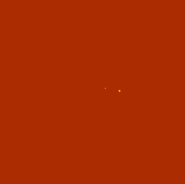}} &
        \subcaptionbox*{Ours}{\includegraphics[width=0.2\columnwidth]{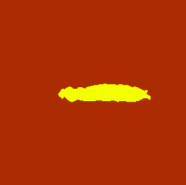}}
    \end{tabular}
    \caption{Visualization of segmentation masks (continued).}
    \label{fig:viz-main-b}
\end{figure}

Furthermore, \figref{fig:viz-main-a} and \figref{fig:viz-main-b} illustrate the segmentation masks generated by our modified architecture as compared to those generated by SimXCA in the hierarchical ViT setting. Despite the noisy groundtruth, our modified architecture is able to generalize quite well as depicted in the top row of the figure. Moreover, our model is also effective in representing classes such as fine sand ripples and small pebbles of rocks, which the vanilla ViT misses in most cases.

\section{Conclusion and Future Work}
\label{sec:conclusion}

In this work we demonstrate the applicability of ViTs for semantic segmentation of the seafloor in SSS waterfalls. To the best of our knowledge, we are the first to employ ViTs for this task. Despite having a small dataset, through our modified design, we achieve results that surpass previous state-of-the-arts by a significant margin while also meeting the computational considerations for real-time implementation.

However, we are still constrained by the lack of precise ground truth to supervise model training. To overcome this weak supervision, we are currently investigating Self-Supervised pre-training followed by Weakly Supervised fine-tuning on image-level labels while also leveraging our noisy ground truth as pseudo masks to regularize training.

Moreover, with the help of the geophysical surveys being conducted by Tecnoambiente SL, we are in the midst of expanding our dataset with additional classes, pixel- and image-level annotations, navigation information and auxiliary metadata. We plan for an eventual release of a large-scale SSS dataset for seafloor segmentation to facilitate further research in this direction.




\printcredits

\section*{Declaration of competing interest}
The authors declare that they have no known competing financial interests or personal relationships that could have appeared to influence the work reported in this paper.

\section*{Acknowledgements}
This study was supported by the DeeperSense project funded by the European Union's Horizon 2020 Research and Innovation programme under grant agreement no. 101016958. We would also like to acknowledge the contribution of Tecnoambiente SL in the dataset generation effort and are grateful for the insightful discussions with Borja Martinez-Clavel on SSS imagery.

\bibliographystyle{cas-model2-names}

\bibliography{bibliography}

\end{document}